%% file: main.tex
\documentclass[conference]{IEEEtran}
\IEEEoverridecommandlockouts

\usepackage{cite}
\usepackage{amsmath,amssymb,amsfonts}
\usepackage{algorithmic}
\usepackage{graphicx}
\usepackage{textcomp}
\usepackage{xcolor}
\def\BibTeX{{\rm B\kern-.05em{\sc i\kern-.025em b}\kern-.08em
    T\kern-.1667em\lower.7ex\hbox{E}\kern-.125emX}}
\usepackage{amsmath,amsfonts}
\usepackage{algorithmic}
\usepackage{algorithm}
\usepackage{array}
\usepackage[caption=false,font=normalsize,labelfont=sf,textfont=sf]{subfig}
\usepackage{textcomp}
\usepackage{stfloats}
\usepackage{url}
\usepackage{verbatim}
\usepackage{graphicx}
\usepackage{cite}

\usepackage[section]{placeins}

\begin{document}
\captionsetup[subfigure]{
 captionskip=0.25pt,
 font=scriptsize,
}

\title{Decentralised, Self-Organising Drone Swarms using Coupled Oscillators}

\author{
\IEEEauthorblockN{Kevin Quinn}
\IEEEauthorblockA{\textit{EEE Department} \\
\textit{Trinity College Dublin}\\
Dublin, Ireland \\
quinnk4@tcd.ie}

\and
\IEEEauthorblockN{Cormac Molloy}
\IEEEauthorblockA{\textit{EEE Department} \\
\textit{Trinity College Dublin}\\
Dublin, Ireland \\
cormac.molloy@tcd.ie}

\and
\IEEEauthorblockN{Harun Šiljak}
\IEEEauthorblockA{\textit{EEE Department} \\
\textit{Trinity College Dublin}\\
Dublin, Ireland \\
harun.siljak@tcd.ie}
}




\maketitle



\begin{abstract}
The problem of robotic synchronisation and coordination is a long-standing one. Combining autonomous, computerised systems with unpredictable real-world conditions can have consequences ranging from poor performance to collisions and damage. This paper proposes using coupled oscillators to create a drone swarm that is decentralised and self organising. This allows for greater flexibility and adaptiveness than a hard-coded swarm, with more resilience and scalability than a centralised system. Our method allows for a variable number of drones to spontaneously form a swarm and react to changing swarm conditions. Additionally, this method includes provisions to prevent communication interference between drones, and signal processing techniques to ensure a smooth and cohesive swarm.

\end{abstract}


\input{Introduction/intro}
\input{Project/sims}
\input{Project/drones}

\input{Project/results}
\input{Conclusion/conc}


\bibliographystyle{IEEEtran}
\bibliography{bib/myrefs}

\end{document}

%% file: Introduction/intro.tex
\section{Introduction}
\IEEEPARstart{C}{oupled}
oscillators are a mathematical model of how spontaneous synchronicity arises in natural and biological systems. Examples include the coordinated pulsing of pacemaker cells in the heart, and the simultaneous flashing of fireflies. One of the most significant early investigations into this phenomenon of spontaneous synchronisation in biological systems was carried out by Mirollo and Strogatz in 1990\cite{STROGATZ1990}. This paper investigated pulse-coupled oscillators, and drew on knowledge from a number of different fields, such as modelling of heart pacemaker cells\cite{PESKIN1975}, analysis of fireflies flashing in unison\cite{BUCK1968}, and earlier work on biological oscillators\cite{WINFREE1967}. 

Strogatz's attempt to model the synchronised flashing of fireflies yielded a model known as the pulse-coupled oscillator. Pulse coupled oscillators are a type of oscillator that communicates with its fellows using discrete pulses. Another example of this type of system in nature is cardiac cells pulsing simultaneously. In general, these types of oscillators make small jumps towards synchronicity by broadcasting pulses of information and reacting to other pulses. In addition to pulse-coupled oscillators, which use discrete-time coupling, there also exist phase-coupled oscillators. In contrast to pulse-coupled oscillators, phase-coupled oscillators interact with their peers continuously\cite{OKEEFFE2017}. A phase-coupled oscillator will typically push or pull the oscillators around it over time to converge gradually toward synchronicity\cite{WINFREE1967}\cite{WIENER1964}. 

The study of swarm behaviour dates back to the 1950s, when equations were derived to describe the movement of schools of fish\cite{BREDER1954}. These described attractive and repulsive forces between aggregations of flocking and swarming animals. Further studies were carried out in the 1980s, when computers became more widely available. In 1982, schooling organisms on a two-dimensional plane were simulated \cite{AOKI1982}. 

The combined study of swarm behaviour and spontaneous synchronisation led O'Keeffe \cite{OKEEFFE2017} to introduce the concept of swarmalators, a portmanteau of 'swarming' and 'oscillators'. This drew on previous research into natural self-organizing systems and combined the element of swarming. Previous research on swarming had largely neglected the study of synchronisation, while studies on synchronisation focused on the internal states of individual agents without regard to their external movement. O'Keeffe et al. examined the collective dynamics of mobile agents with coupled internal phase and spatial dynamics. 

While swarming is a famous challenge and use case for unmanned aerial vehicles (UAVs), issue of synchronisation for drones and the equipment for communications and sensing they carry is a significant open challenge \cite{podbregar2024feasibility}. The concept of multi-drone synchronisation using pulse-coupled oscillators has already been investigated by Breza \cite{BREZA2023}, but this demonstration stops short at synchronisation, with stationary drones. Barcis \cite{BARCIS2020} implemented a synchronising and swarming model on real-world hardware, primarily focused on a two-dimensional implementation of swarmalators and using wheeled Pololu Balboa robots. The model was demonstrated on a Crazyflie quadcopter platform, but lacked decentralisation, using a base station for control. However, these experiments prove the feasibility of a real-world swarmalator on a physical platform. 

This paper proposes a model which allows for both decentralised synchronisation and swarming, with drones which can self-organise into predictable structures. We will show how the coupled oscillator and swarmalator concepts can be applied to multi-drone systems. This solution is more flexible than a pre-programmed swarm and eliminates the single point of failure of a single base station. It also demonstrates the capability of self-synchronisation in groups of autonomous robots.





%% file: Project/sims.tex
\section{Initial Investigation}
\subsection{Pulse-Coupled Oscillator Simulation}
As an initial exploratory exercise into the properties of coupled oscillator systems, we first develop a simulation of a pulse-coupled oscillator system as described in \cite{STROGATZ1990}. The working principles of this type of oscillator are:
\begin{enumerate}
  \item Individual agents have an internal phase in the range of 0 - 1.
  \item Agents are not able to see other agents' phases.
  \item Each agent's phase increases at a constant rate from 0 to 1.
  \item When an agent's phase reaches 1, the phase resets to 0 and a 'pulse' is released.
  \item Other agents are able to see these pulses and adjust their own internal phase accordingly. 
\end{enumerate}

\begin{figure}[!t]
\centering
\subfloat[]{\includegraphics[width=3.15in]{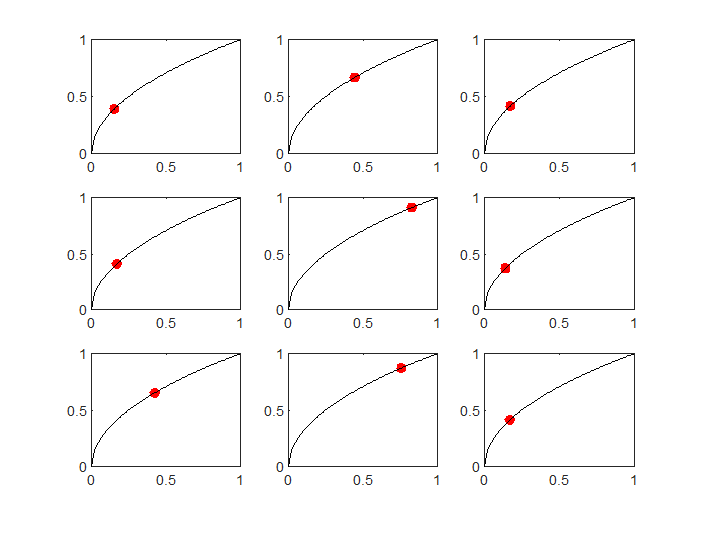}%
\label{PulseSimInit}}

\subfloat[]{\includegraphics[width=3.15in]{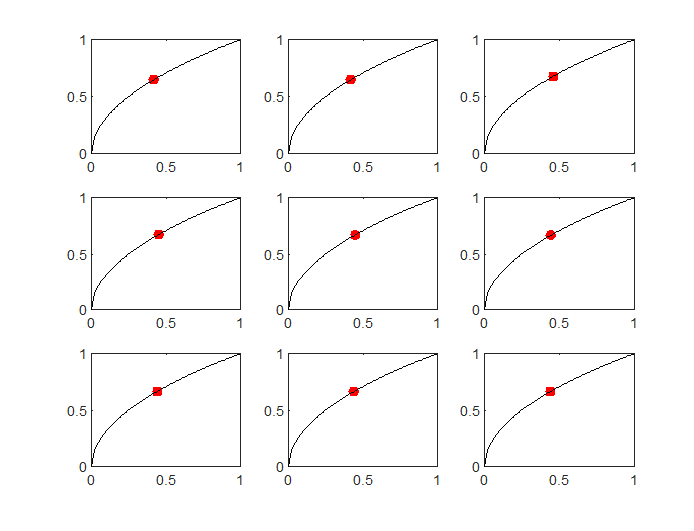}%
\label{PulseSimEnd}}
\caption[Pulse-coupled simulation]{The pulse-coupled oscillator simulation with $N_o$=9 oscillators. Figure (a) shows the starting conditions with each oscillator randomised. Figure (b) shows the oscillators shortly after, in synchronisation.}
\label{fig_sim}
\end{figure}

The oscillator phases are visualised in Figs. \ref{PulseSimInit} and \ref{PulseSimEnd} as red dots moving up a curve of equation $y = \sqrt{x}$. One plot is drawn for each oscillator. 

The equation governing the response of an individual oscillator to a pulse is given by:
\begin{equation}
\label{eqn:pulseCouple}
\theta_{new} = (\sqrt{\theta} + K_C)^2
\end{equation}
where $\theta$ is the oscillator's internal phase (0 - 1) and $K_C$ is the coupling constant \cite{STROGATZ1990}. $K_C$ must be chosen carefully, as an overly large coupling constant can result in the oscillators making large jumps in phase, resulting in instability, while a very low coupling constant will result in the oscillators requiring a large amount of time to synchronise. 

For illustration, Figure \ref{PulseSimInit} shows the initial conditions of the pulse-coupled oscillator simulation with $N_o=9$ oscillators. The x-axis represents phase and the y-axis represents responsiveness to another pulse. When an oscillator reaches the end ($x=1$) of its graph, it sends a pulse. When an oscillator receives a pulse, its phase is advanced in proportion to its y-coordinate according to (\ref{eqn:pulseCouple}). Fig. \ref{PulseSimEnd} shows the end conditions of the simulation, where each oscillator has an approximately equal phase. This simulation therefore demonstrates capability for spontaneous synchronisation between multiple independent agents.

\begin{figure}[!t]
\centering
\subfloat[]{\includegraphics[width=3in]{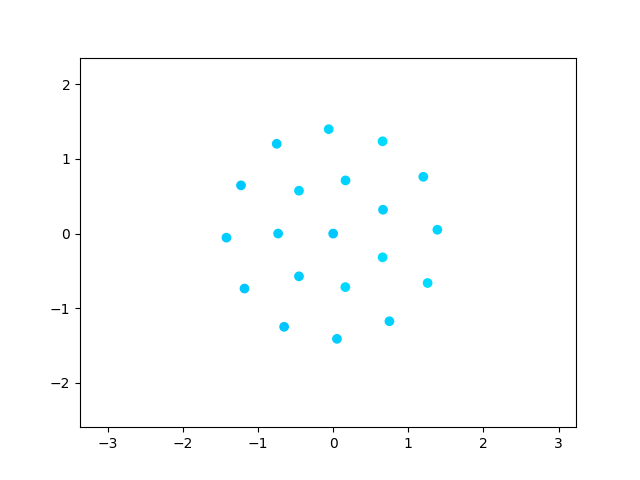}%
\label{SwarmalatorSync}}
\\
\subfloat[]{\includegraphics[width=3in]{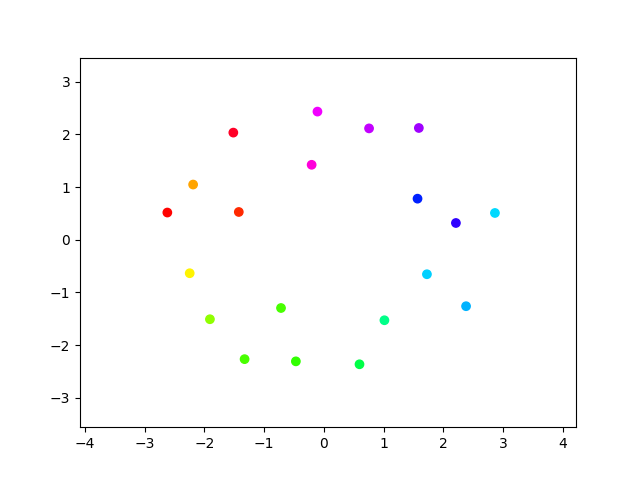}%
\label{SwarmalatorDesync}}
\caption[Swarmalator simulation]{The Python swarmalator simulation ($N_o$ = 20) with a positive coupling coefficient ($K_C$ = 0.7) in figure (a) and a negative coupling coefficient ($K_C$ = -0.7) in figure (b).}
\label{SwarmalatorSim}
\end{figure}

\subsection{Swarmalator Simulation}
Now we create a  swarmalator simulation derived from the O'Keeffe\cite{OKEEFFE2017} model. The rules for this type of oscillator are as follows:
\begin{enumerate}
  \item Individual agents have an internal phase $\theta$ in the range of 0 - 2$\pi$.
  \item Agents are able to see other agents' phases.
  \item Each agent's phase increases at a constant rate from 0 to 2$\pi$.
  \item When an agent's phase reaches 2$\pi$, the phase wraps around to 0.
  \item Agents are able to adjust their own internal phase according to the phases of the other oscillators in the system. 
  \item Individual agents have a unique location (X and Y).
  \item Agents are able to see other agents' location.
  \item Agents have a short-range repulsive force which prevents them from occupying the same space.
  \item Agents have a long-range attractive force which is stronger the more similar their phases.
  \item The movement of individual agents is strictly governed by the short-range repulsive and long-range attractive forces. 
\end{enumerate}

This is more analogous to a phase-coupled oscillator system than a pulse-coupled one, as there is continuous communication between agents. However, it serves as a useful starting platform for the investigation of pulse-coupled swarmalators. 

O'Keeffe's equation governing the positioning of an individual oscillator is given by:
\begin{equation}
\label{eqn:swarmMove}
\mathbf{x_{inew}} = \frac{1}{N_o} \sum_{j \neq i}^{N_o}\frac{\mathbf{x_j - x_i}}{\mathbf{|x_j - x_i|}}(A + J\cos(\theta_j - \theta_i)) - B\frac{\mathbf{x_j - x_i}}{\mathbf{|x_j - x_i|}^2} 
\end{equation}
where $\mathbf{x} = [x\, y]$ is the X and Y coordinate of the oscillator, $A$ is long-range attraction, $J$ is the \emph{like-attract-like} strength, and $B$ is the short-range repulsion strength.

O'Keeffe's equation governing the phase response of an individual oscillator to the collective phase is given by:
\begin{equation}
\label{eqn:swarmCouple}
\theta_{inew} = \theta_i + \omega_i + \frac{K_C}{N_o} \sum_{j \neq i}^{N_o}\frac{\sin(\theta_j - \theta_i)}{|\mathbf{x_j - x_i}|} 
\end{equation}
where $\omega$ is the frequency of each oscillator, and $N_o$ is the number of oscillators \cite{OKEEFFE2017}. 

Additionally, the base frequency $\omega$ of each agent is separately modified by a frequency variation parameter. The addition of this parameter allows the modelling of dissimilar oscillators - for example, two agents (drones, robots, etc.) with slightly different internal clocks.

Two simulations were run, the first demonstrating positive coupling and the second negative. These demonstrated the expected behaviour of an ideal swarmalator under both conditions. The parameter values used for these simulations are given in Tables \ref{tab:SwarmSyncParams} and \ref{tab:SwarmDesyncParams}.

\begin{table}
\parbox{.45\linewidth}{
\caption{Swarmalator Synchronisation Values\label{tab:SwarmSyncParams}}
\centering
\begin{tabular}{|c||c|}
\hline
\textbf{Parameter} & \textbf{Value}\\
\hline
$N_o$ & 20 \\
$K_C$ & 0.7 \\
$J$ & 0.8 \\
$B$ & 3 \\
$A$ & 1 \\
\hline
\end{tabular}
}
\hfill
\parbox{.45\linewidth}{
\caption{Swarmalator Desynchronisation Values\label{tab:SwarmDesyncParams}}
\centering
\begin{tabular}{|c||c|}
\hline
\textbf{Parameter} & \textbf{Value}\\
\hline
$N_o$ & 20 \\
$K_C$ & -0.7 \\
$J$ & 0.8 \\
$B$ & 3 \\
$A$ & 1 \\
\hline
\end{tabular}
}
\end{table}

At the beginning of the simulation, the swarmalators are disorganised and unsynchronised. However, they quickly form a structure. Fig. \ref{SwarmalatorSync} shows the swarmalators in a ``static sync'' pattern - the swarmalators are unmoving and approximately equidistant, and their phases are synchronised. The like-attracts-like force evenly balances the repulsive force and keeps them in a disc formation, while the phases pull each other towards synchronisation and keep them there. This is accomplished by using a positive coupling coefficient, little to no frequency variation, and a positive like-attract-like and repulsion value.

A qualitatively different behaviour emerges when the simulation is given a negative coupling coefficient (Table \ref{tab:SwarmDesyncParams}). Oscillators push past and move around each other chaotically. Each oscillator naturally attracts other oscillators of the same phase, which then exert influence on changing each other's phases, and once their phases have changed they then repel each other. The order that emerges is a population of oscillators with phases distributed relatively evenly across the entire range. These oscillators physically attract those of closest phase, resulting in the rainbow effect seen in Fig. \ref{SwarmalatorDesync}. 


The findings from these experiments are as follows:
\begin{enumerate}
  \item Both pulse- and phase-coupled oscillators are capable of rapid synchronisation and desynchronisation.
  \item A continuous-communication system can be emulated with a pulsed-communication system, by driving up the rate at which pulses are sent until time between pulses is negligible.
  \item The more oscillators there are on a given phase, the more attracted other oscillators will be to that phase.
  \item Swarmalators can form complex structures and arrangements by following simple rules. 
  \item Swarmalator behaviour can be modified and directed by modification of their parameters.  
\end{enumerate}

%% file: Project/drones.tex
\section{Drone Syncing and Swarming}
\subsection{Pulse-Coupled Swarmalators}
Following successful simulation, we develop a novel model for use with the Crazyflie drone. This new model introduced the limitation that all inter-agent communication needed to be pulse-based, rather than continuous, in order to avoid message interference. This also reduced the computational load on each individual drone. 

The O'Keeffe \cite{OKEEFFE2017} swarmalator model was chosen as a starting point. However, as that model was based on continuous communication and coupling, it needed to be modified to utilise discrete, pulse-based communication. Recalling (\ref{eqn:swarmCouple}), it can be seen that the new phase is calculated from the sum of all other phases. Additionally, as the phases are constantly changing, all phases involved in the calculation must be from a single instant. However, due to message interference, all oscillators cannot share their phases at the same time. They must be shared one at a time. Therefore, either the phases of each oscillator must be collected one by one and their values at moment of calculation extrapolated from the stored values, or the value of $\theta_{inew}$ must be recalculated at each instant that another oscillator's phase is received. The latter option was chosen, due to concerns over the accuracy of extrapolated values and computational complexity. The new phase coupling equation is given by:
\begin{equation}
\label{eqn:dronePhaseCouple}
\theta_{inew} = \theta_i + K_C \cdot \sin(\theta_j - \theta_i) 
\end{equation}

The swarming equation (\ref{eqn:swarmMove}) was modified to:
\begin{equation}
\label{eqn:droneMove}
\mathbf{x_{inew}} = \frac{\mathbf{x_j - x_i}}{|\mathbf{x_j - x_i}|}(A + J\cos(\theta_j - \theta_i)) - B\frac{\mathbf{x_j - x_i}}{|\mathbf{x_j - x_i}|^2} 
\end{equation}
These equations resemble their simulation-based counterparts, but exclude the summation and division over all agents in the swarm. A summary of the modes of operation of each type of oscillator can be seen in Table \ref{tab:types}.

\begin{table}[h]
\caption{Comparison between differing types of oscillators\label{tab:types}}
\centering
\begin{tabular}{|c||c||c|}
\hline
\textbf{Type} & \textbf{Coordination} & \textbf{Communication}\\
\hline
Pulse-coupled       & Temporal                  & Pulsed \\
\hline
Phase-coupled       & Temporal                  & Continuous \\ 
\hline
Swarmalator         & Temporal and spatial      & Continuous \\ 
\hline
Drone-based model   & Temporal and spatial      & Pulsed \\ 
\hline
\end{tabular}
\end{table}

\subsection{Smoothing}
\begin{figure}[t!]
      \centering
      \includegraphics[width=0.4\textwidth]{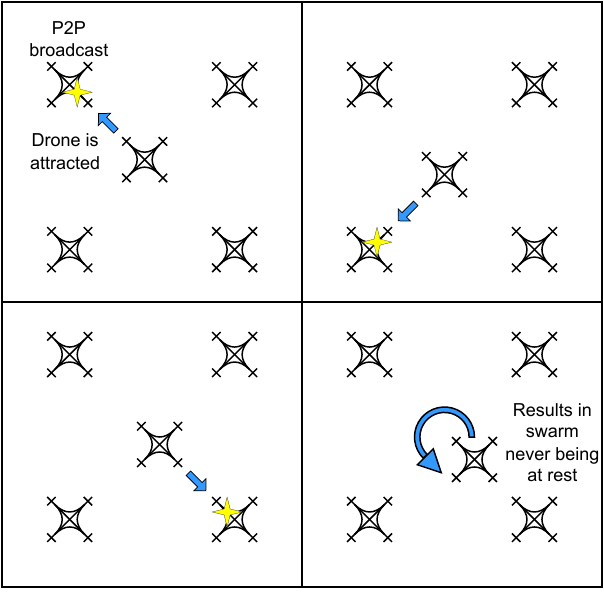}
      \caption[Swarm Diagram]{A visualisation of the type of movement visible within the swarm when no smoothing or averaging is applied.}
      \label{SwarmDiagram}
\end{figure}

These equations assume a two-drone system, between the receiver and the most recent other drone to have broadcast. The two-drone system is then redefined each time a new drone broadcasts---one drone broadcasts, the other \textit{N-1} drones receive and independently recalculate their own movements. The benefit of this adaptation is the lack of requirement for each drone to know how many others are in the swarm--- reducing the amount of data required to be shared between drones, and allowing drones to continuously enter and leave the swarm if needed. The drawback of this approach is that at any given instant, the drone is solely influenced by the most recent member to have transmitted, rather than being influenced by the entire swarm. The effects of this are visible as small, shuffling movements within the swarm. This can be seen in Fig. \ref{SwarmDiagram}.

To address this issue, two methods were tested. The first was to use a moving average function to average the effects of the previous $N$ peer to peer (P2P) broadcasts, where $N$ is a variable chosen as a medium between stability and responsiveness. The formula for this type of smoothing is given by (\ref{eqn:movingAvg}). 

\begin{equation}
\label{eqn:movingAvg}
\hat X[n] = \sum_{i=0}^{N-1} \frac{X[n-i]}{N}
\end{equation}

The second method trialled was to use an exponential smoothing function, where the previous values in combination with a parameter $\alpha$ are used to determine a smoothed value. The formula for this type of smoothing is given by (\ref{eqn:expSmoothing}).
\begin{equation}
\label{eqn:expSmoothing}
\hat X[n] = (\alpha \cdot X[n]) + ((1 - \alpha) \cdot \hat X[n-1])
\end{equation}


\subsection{Dual-Phase Oscillators}
In the process of testing drone-to-drone synchronisation, the issue of interference arose. It was possible for multiple drones to send P2P pulses at the same time, resulting in overlapping messages and poor overall performance. In order to preclude this possibility, the drones needed a way to stagger their communications while synchronising their phases. To do this, inspiration was drawn from the simulations of negatively-coupled oscillators. Each drone was given a second, hidden phase, in addition to its primary phase. The goal of the hidden phase was to regulate communication between agents - while the primary phase was driven to either synchronisation or desynchronisation, the hidden phase would always be driven to desynchronisation. This resulted in the hidden phase of each oscillator being unique, which ensured evenly spaced P2P communications between drones.

%% file: Project/results.tex
\section{Results}
\input{Project/setup}
\subsection{Synchronisation}

\begin{figure}[!h]
\centering
\subfloat[]{\includegraphics[width=0.45\textwidth]{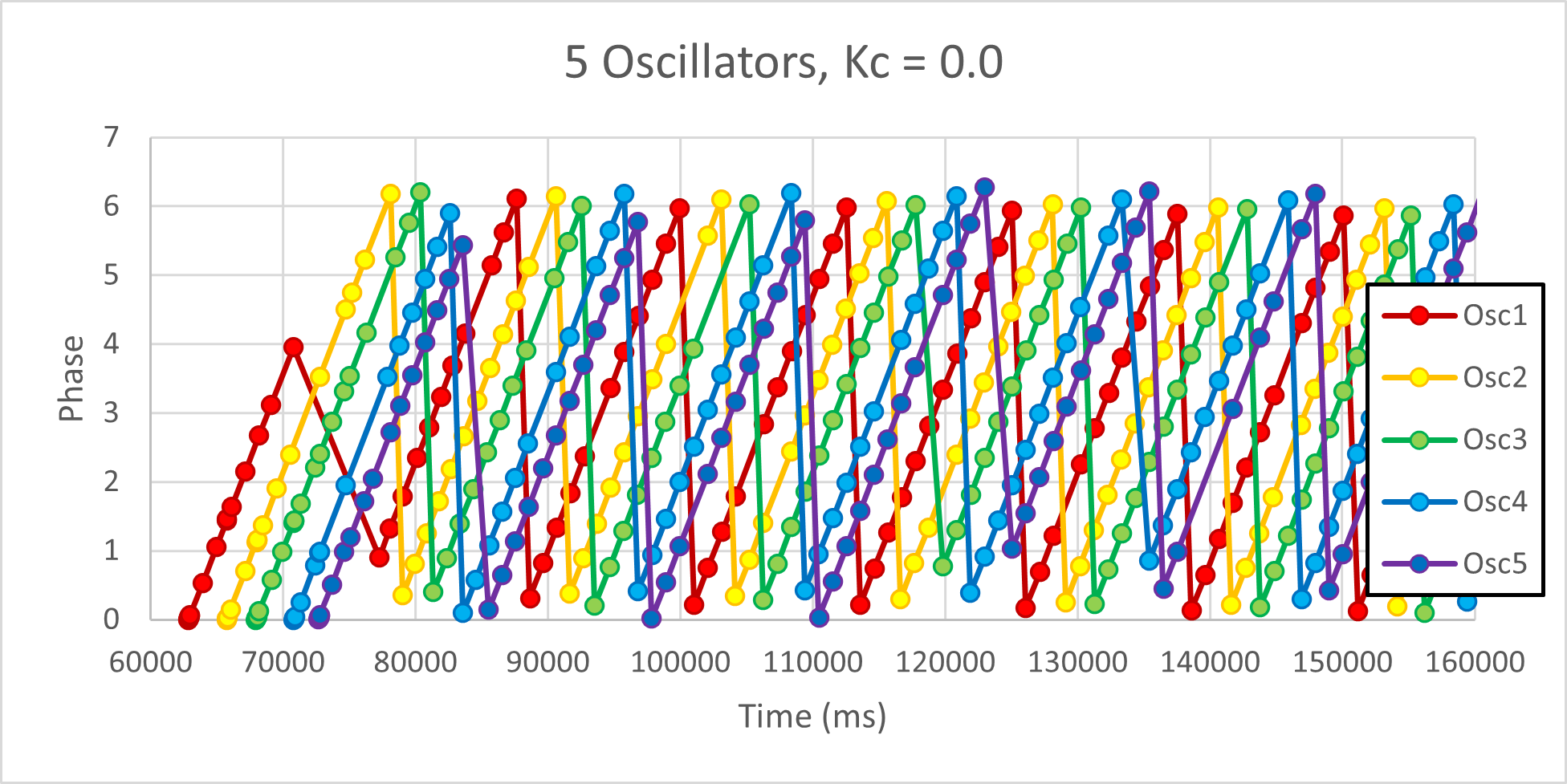}
\label{Sync000}}


\subfloat[]{\includegraphics[width=0.45\textwidth]{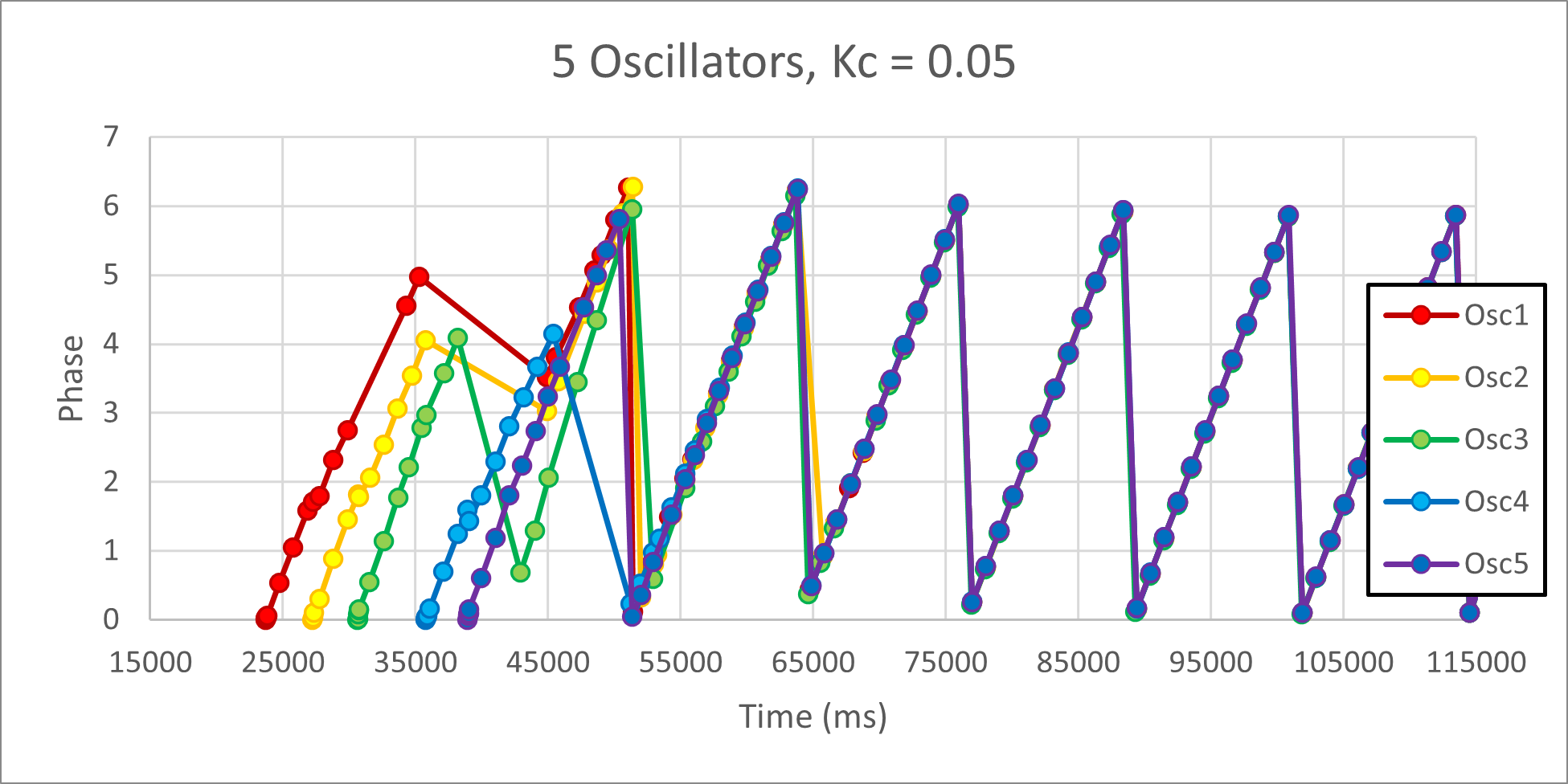}
\label{Sync005}}


\subfloat[]{\includegraphics[width=0.45\textwidth]{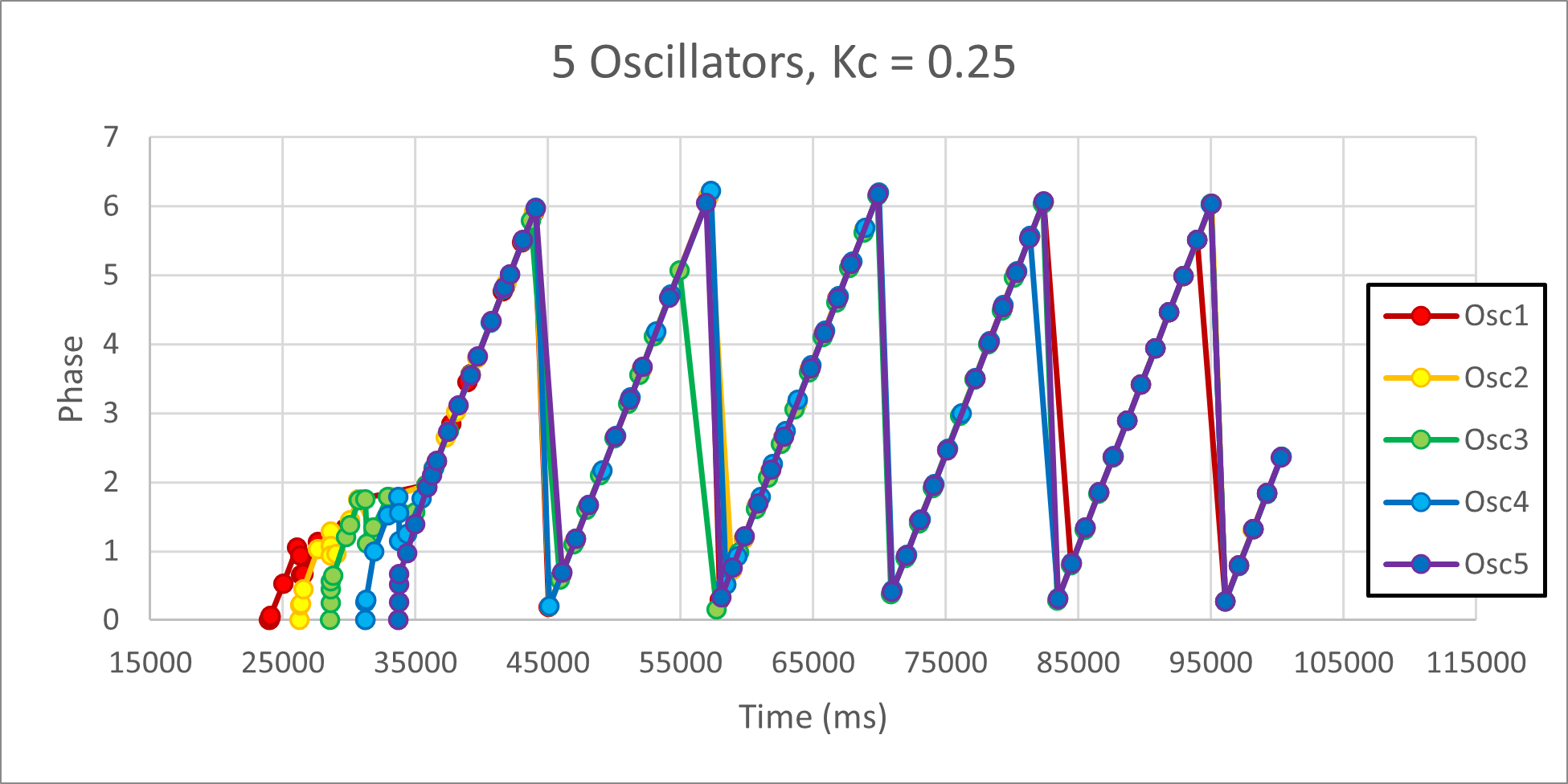}
\label{Sync025}}
\caption[Synchronisation Graphs]{A series of graphs comparing the effects of different values of coupling constant $K_c$ on rate of coupling.}
\label{Sync}
\end{figure}

Synchronisation between drones worked well, with groups of drones taking less than a couple of seconds to synchronise. This can be seen in Fig. \ref{Sync}, where the phase of 5 drones is tracked with various coupling coefficients. With no coupling (Fig. \ref{Sync000}), drones neither converge nor diverge, while with higher values (Fig. \ref{Sync005}) coupling happens more promptly. Exceedingly high values, as seen in Fig. \ref{Sync025} can cause negative effects - it can be seen that when a new oscillator is added to the system, all existing oscillators are immediately pulled to the new oscillator's phase, rather than mutual convergence to the same phase.

Additionally, removing a drone from the swarm and adding it back with a different phase did not prove disruptive to the overall synchronisation of the swarm, with the outlier drone quickly adjusting and merging phase with the rest of the swarm. Tests were also run with the drones oscillating with slightly different frequencies, which synchronised quickly given a sufficiently high coupling coefficient and low frequency variation. This demonstrates that the algorithm is both efficient, leading to quick synchronisation, and robust, quickly adapting to and subsuming new drones or dissimilar ones.

\subsection{Swarming}
Testing on the swarming algorithm was performed by placing up to five drones in a cross, or quincunx, shape in the testing area. They then lifted off together before beginning swarming. Pulses from each drone were monitored to build up a picture of how the swarm behaved. Tests were carried out with no smoothing, with exponential smoothing and with moving average smoothing.

\begin{figure}[!t]
\centering
\subfloat[]{\includegraphics[width=0.24\textwidth]{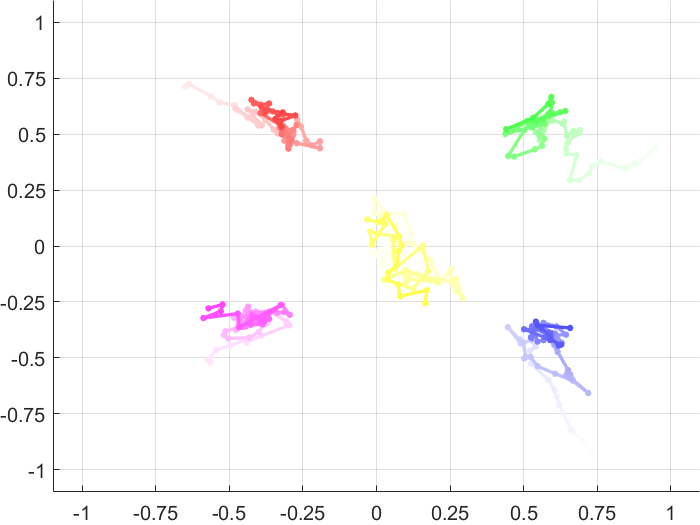}%
\label{NoSmooth}}%
\hfil
\subfloat[]{\includegraphics[width=0.24\textwidth]{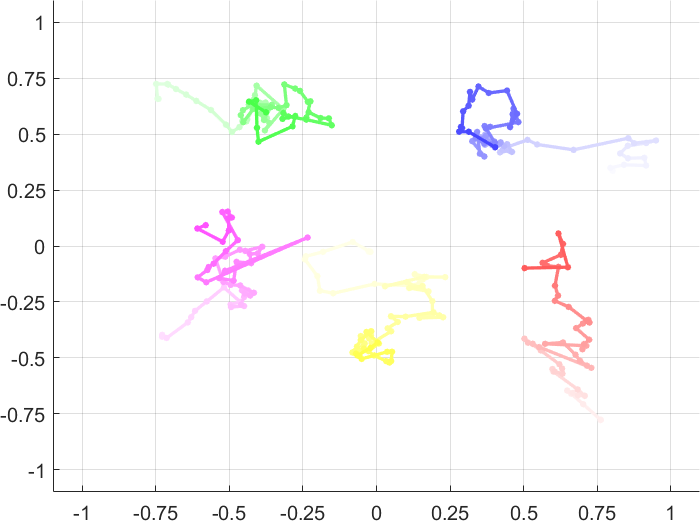}
\label{ExpSmoothAlph08}}

\subfloat[]{\includegraphics[width=0.24\textwidth]{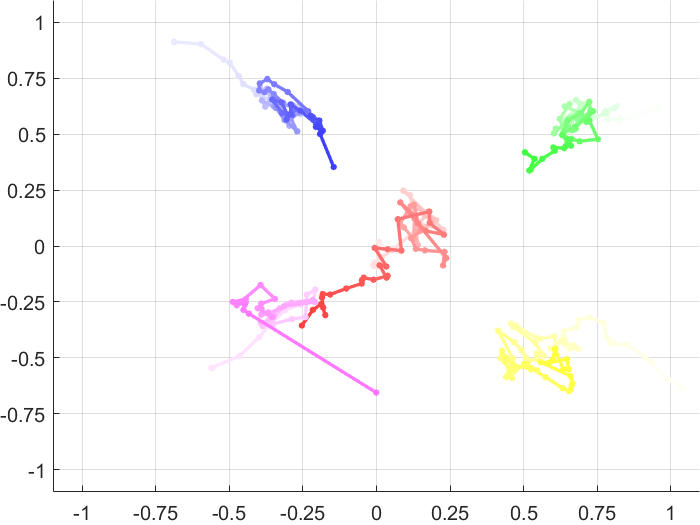}%
\label{MovAvg10}}%
\hfil
\subfloat[]{\includegraphics[width=0.24\textwidth]{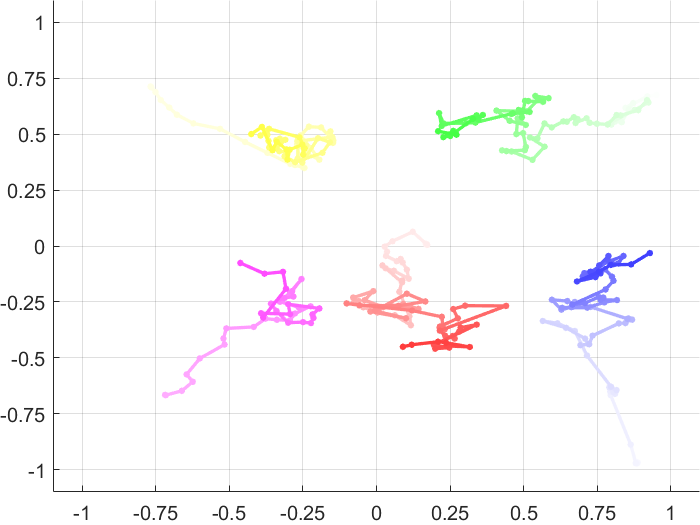}%
\label{MovAvg20}}%
\caption[Swarm Paths]{The paths taken by drones in swarming tests. More recent data represented by higher colour saturation. Upper left - no smoothing, upper right - exponential smoothing ($\alpha$ = 0.8), lower left - moving average ($N$ = 10), lower right - moving average ($N$ = 20).}
\label{SwarmPaths}
\end{figure}

For the first test, no smoothing algorithm was applied. This demonstrated relative movement of each drone towards each other (Fig. \ref{NoSmooth}). It can be seen that the four outermost drones demonstrate reciprocal motion to and from the centre of the swarm, while the central drone is shuffled around as described in Fig. \ref{SwarmDiagram}. It can also be seen that the drones remain in the quincunx formation in which they started - in simulation, five ideal swarmalators will move to form a pentagon in order to minimise the distance between them.


\begin{figure}[!t]
\centering
\subfloat[]{\includegraphics[width=0.24\textwidth]{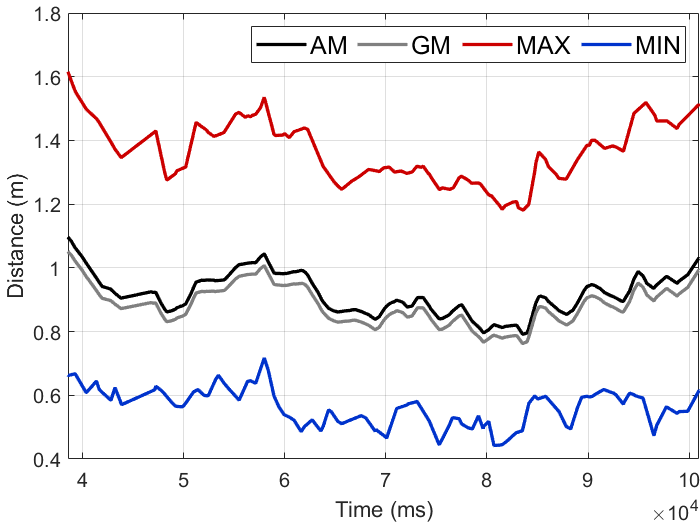}%
\label{MeanDistNoSmooth}}%
\hfil
\subfloat[]{\includegraphics[width=0.24\textwidth]{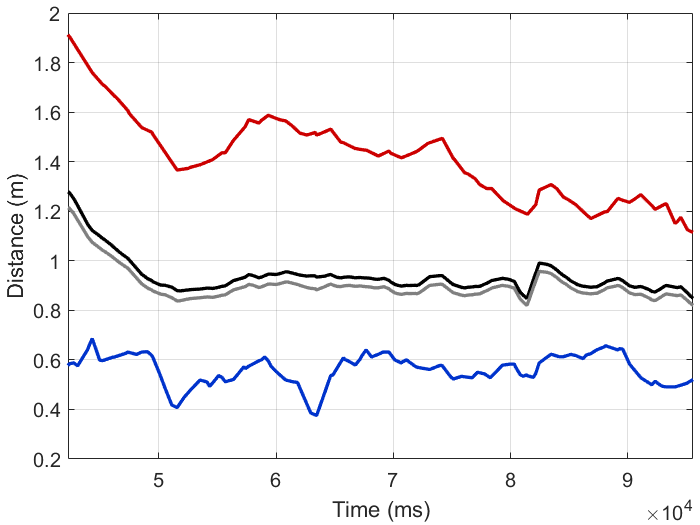}%
\label{MeanDistExpSmoothAlph08}}

\subfloat[]{\includegraphics[width=0.24\textwidth]{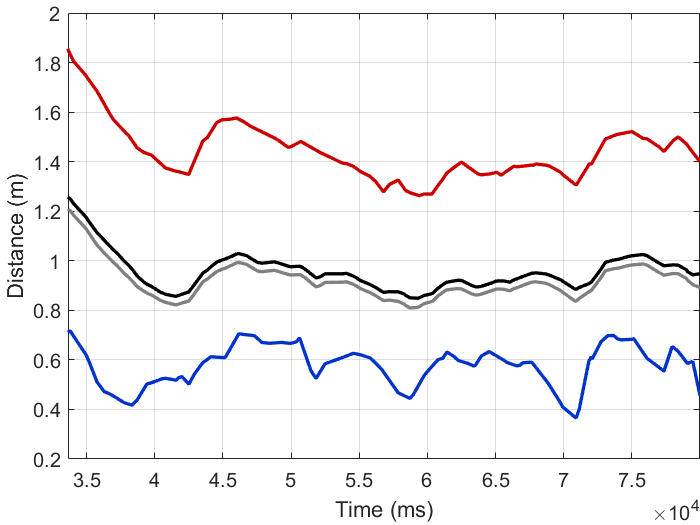}%
\label{MeanDist_MovAvg10}}%
\hfil
\subfloat[]{\includegraphics[width=0.24\textwidth]{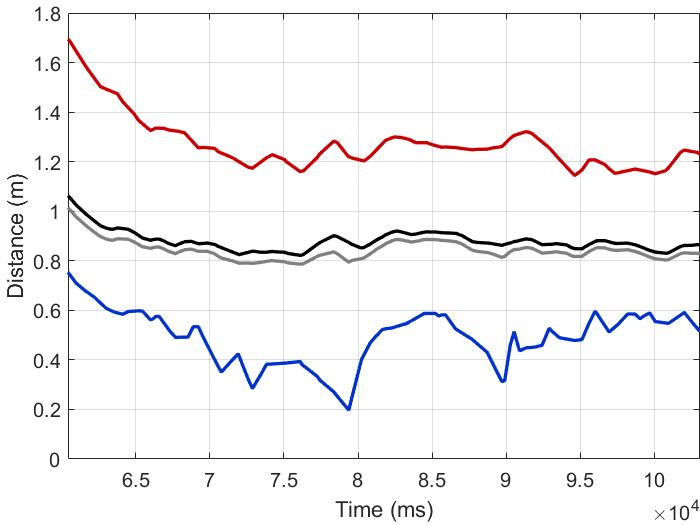}%
\label{MeanDist_MovAvg20}}%
\caption[Mean Distances]{The arithmetic (AM) and geometric mean (GM) distances between drones in swarming tests. Upper left - no smoothing, upper right - exponential smoothing ($\alpha$ = 0.8), lower left - moving average ($N$ = 10), lower right - moving average ($N$ = 20).}
\label{MeanDist}
\end{figure}

The second test, with exponential smoothing gave improved results. Testing was done with a range of values of $\alpha$, with values of 0.2 and 0.8 giving best results. Rather than remaining in approximately the starting formation, the drones moved to form the pentagon shape as seen in simulation (Fig. \ref{ExpSmoothAlph08}). To help draw a comparison to the tests done with no smoothing, the mean distance between drones over time was calculated. This would clearly show how closely the drones can fly in formation, as well as how much movement exists between them. The geometric mean and arithmetic mean distances between exponentially smoothed drones can be seen in Fig. \ref{MeanDistExpSmoothAlph08}. The mean distance initially sharply decreases, then levels off as the drones assume the formation. The exponential smoothing then shows that mean distance remains quite stable, with maximum and minimum slowly converging together (with a pentagonal formation, it is impossible for maximum and minimum to fully equal). In comparison, the experiment with no smoothing gives a very rough and unstable mean distance between drones (Fig. \ref{MeanDistNoSmooth}). This is because the drones are moving in a reciprocal manner towards and away from each other. The minimum and maximum values are similarly unstable and do not converge.

The third set of tests was done with a moving average smoothing technique. This was again done on a range of values of $N$, with 10 and 20 giving the best results. The drones are again seen to assume the pentagonal formation. Note that with the $N=10$ simulation (Fig. \ref{MovAvg10}), the drone in magenta experienced a failure halfway through the test, and the red, green, and blue drones moved to close the formation in its place. This demonstrates the adaptability and responsiveness. 

A mean distance between drones was once again calculated for each of these tests\footnote{excluding the second half of $N=10$}. It is clear that the lower depth of moving average results in a quicker response time, as the mean distance decreases more rapidly, but after levelling off it is not quite as steady as the $N=20$ test. This gives a clear picture of the tradeoff between responsiveness and stability that is posed by using a moving average. Additionally, it was observed that with smoothing algorithms in place (both exponential and moving average) the drones flew much more steadily than with no smoothing, leading to fewer instances of drones losing stability and falling. This shows the importance of the smoothing algorithms, but also demonstrates a key point: real, physical drones or swarmalators are fundamentally different to simulated, ideal swarmalators, as the movement of an ideal swarmalator is strictly governed by its equations of movement. It is not subject to any forces other than the direct attractive or repulsive forces of its neighbours. Real-world swarmalators have other forces to contend with, such as air turbulence, vibration, and other outside effects.

%% file: Project/setup.tex

The platform  used for the experiment was the Bitcraze Crazyflie 2.1\footnote{https://www.bitcraze.io/products/crazyflie-2-1/ - accessed 09/04/2023}. This was used in conjunction with the HTC Vive Lighthouse station which provided positioning data. Four receivers on the drone received infrared light from the Lighthouse base stations and Kalman Filtering was implemented on board to give an estimation of drone state.

%% file: Conclusion/conc.tex
\section{Conclusion}
This paper proposes a biologically-inspired method of syncing and swarming robots based on the coupled oscillators concept. The experiments performed showed the capabilities of this method. Some of the more interesting of these include rapid synchronisation between independent and dissimilar agents, a robust swarming method responsive to changes in the swarms constitution, and the capability for autonomous operation without ground-based input or control. 

This method quickly synchronises independent agents with short pulses, and, given a sufficiently well-chosen coupling constant, can synchronise agents of differing fundamental frequencies. It is robust to agents leaving or dropping out of the swarm, and will rapidly reshape and reform to cover up any gaps left by those agents. This is done automatically, without human input. Synchronisation is not affected when this happens. Our approach is permissive of agents entering the swarm at any time. The swarm will reform and adapt to any new agents entering, and hypothetically could allow for two swarms to merge, given sufficient synchronisation strength. These drones will act identically to original members of the swarm, and a drone entering the swarm after its formation is no different to one present at the outset.

In our future work, we will look at scaling challenges of the proposed method, and applications of synchronisation in communication and sensing equipment on board, in the context of 6G non-terrestrial networks.